\documentclass{article}

\usepackage[final]{neurips_2022}

\usepackage{enumitem}

\usepackage[utf8]{inputenc} 
\usepackage[T1]{fontenc}    
\usepackage{hyperref}       
\usepackage{url}            
\usepackage{booktabs}       
\usepackage{amsfonts}       
\usepackage{nicefrac}       
\usepackage{microtype}      
\usepackage{xcolor}         

\hypersetup{
   colorlinks=true,
   linkcolor=[RGB]{30, 30, 180},
   citecolor=[RGB]{30, 30, 180},
   urlcolor=magenta,
   pdfborder=0 0 0,
   pdftitle={},
   pdfsubject={}, 
   pdfkeywords={},
   pdfauthor={},%
   pdfstartview=FitH
}

\defcitealias{ArtificialIntelligenceForEurope}{COM(2018) 237 final}
\defcitealias{EUAIAct21}{COM(2021) 206 final}

\title{Functional trustworthiness of AI systems by statistically valid testing\\ \vspace{0.5cm} \normalsize Position paper to the current regulation and standardization effort of AI in Europe}

\author{%
  Bernhard Nessler \\	
  Research Manager Deep Learning \& Certification \\
  Software Competence Center Hagenberg \\
  \texttt{bernhard.nessler@scch.at} \\
  \And
  Sepp Hochreiter \\
  Institute for Machine Learning\\
  Johannes Kepler University Linz \\
  \texttt{hochreit@ml.jku.at} \\
  \And
  Thomas Doms \\
  Global Product Lead AI Services \\
  TÜV AUSTRIA Holding \\
  \texttt{thomas.doms@tuv.at} \\
}

\begin{document}

\maketitle

\begin{center}
    {\large \bf Executive Summary}
\end{center}

The authors are \textbf{concerned about the safety, health, and rights of the European citizens} due to inadequate measures and procedures required by the current draft of the EU Artificial Intelligence (AI) Act for the conformity assessment of the functional accuracy, robustness, and transparency of AI systems. We observe that not only the current draft of the EU AI Act, but also the accompanying standardization efforts in CEN/CENELEC, have resorted to the position that real functional guarantees of AI systems supposedly would be unrealistic and too complex anyways. Yet enacting a conformity assessment procedure that creates the false illusion of trust in insufficiently assessed AI systems is at best naive and at worst grossly negligent. 

The EU AI Act regulates that a huge volume of documentation has to be written, but it grossly fails to define any level of \textbf{testable quality requirements} for future automated decision. In order to recognize the shortcomings, one has to know that the difference between a trustworthy professionally-developed AI decision system and a non-trustworthy carelessly assembled AI decision system lies exclusively in \textbf{the precision of the definition of its application domain and the fact that it was statistically tested \footnote{Note that we refer to high-dimensional, data-driven AI methods for which full functional proofs can´t be given} on that very domain}. It should be well-known, that in contrast to classical engineering techniques, AI, or more precisely deep machine learning (ML/DL) techniques\footnote{due to the pseudo-chaotic nature of the high-dimensional optimization process}, suffer from the possibility of a bad generalization capability that is often unexpected by the user. 

This is as if we would create a knife. Its \textit{intended purpose} is cutting vegetables. Its cutting performance was trained on some apples, green broccoli, ripe watermelons, and some more vegetables. But upon using it, we realize that it does not at all cut through "Belle de Boskoop" apples, squeezes non-strait cucumbers instead of cutting them and has problems with some unripe melons. And astonishingly, it fails most of the time if we try using it anywhere outside of the kitchen.   
This generalization problem is deeply inherent to AI/ML. This is why there are well-known solid mathematical techniques, developed more than 40 years ago by the ML-community, that clarify that the generalization capability of the ML model has to be \textbf{empirically measured with statistically valid random tests}, in the same way as we measure the efficacy of modern medical drugs in randomized studies today. 
For the knife, we would probably have wanted to ask, \textbf{what is the probability that the knife fails} to cut a vegetable while preparing a meal in a non-commercial kitchen in any private household in Europe. And note that using the same knife in a commercial kitchen or in a private household in China would likely yield different probabilities of failing.
Now imagine that the product is not just a single knife, but a fully automated kitchen robot that could potentially harm the user.

The trustworthiness of an AI decision system lies first and foremost in the correct statistical testing on \textbf{randomly selected samples} and in the precision of the definition of the application domain, which enables drawing samples in the first place. We will subsequently call this testable quality \textit{functional trustworthiness}. It includes a design, development, and deployment that enables correct statistical testing of all relevant functions. All other requirements and further inspection methods and security measures, like human oversight, logging obligations, and guards against  adversarial or otherwise malicious attacks\footnote{like e.g. \citep{greshake2023youve}}, are also important for trustworthy AI systems, but they come second to the trustworthiness of the statistical quality of the function itself.

The EU AI Act thus misses the point of ensuring quality by functional trustworthiness and correctly attributing responsibilities. Just asking about the \textit{intended purpose} is supporting the careless style of badly engineered and insufficiently tested ML hacks. The term \textit{sufficiently representative} is meaningless for a test dataset, if it wasn't sampled truly at random from a well-defined distribution \textbf{after the model was trained}\footnote{This essential random character of statistical testing is never mentioned in Article 10, nor elsewhere in the AI Act.}. Similarly the requirement of \textit{unbiased training datasets} does not help the cause as long as there are no goal definitions for non-discrimination that could be statistically tested.

We are firmly convinced and advocate that a reliable assessment of the statistical functional properties of an AI system according to the well-known scientific state of the art has to be the indispensable, mandatory nucleus of the conformity assessment. In this paper, we describe the \textbf{three necessary elements} to establish a reliable functional trustworthiness, i.e., (1) the \textbf{definition of the technical distribution of the application}, (2) the risk-based \textbf{minimum performance requirements}, and (3) the \textbf{statistically valid testing} based on independent random samples.

\section{Introduction}
Modern artificial intelligence (AI) is one of the fastest growing technologies of the 21\textsuperscript{st} century and accompanies us in our daily lives when interacting with technical applications. 
In the EU economic area, this innovative technology is seen as a key and success factor for future economic prosperity.

\paragraph*{Problems in practice.}
Due to the increased availability of large amounts of data and better hardware, the field of AI, and in particular machine learning (ML) and deep learning (DL), the driving forces of modern AI, have experienced fast developments and created substantial scientific discoveries during recent years. 
This fast progress of scientific discoveries and the pressure to immediately monetize them in practical applications are prone to increasing the risk of reduced quality of the resulting AI systems. 

\paragraph*{Current legal status in the EU.}
However, reliance on such technical systems is crucial for their widespread applicability and acceptance. The societal tools to express reliance are usually formalized by lawful regulations, i.e., standards, norms, accreditations, and certificates. 
In addition, the EU's AI strategy is laid out explicitly in its AI governance. It states that ``the EU must therefore ensure that AI is developed and applied in an appropriate framework which promotes innovation and respects Union’s values and fundamental rights as well as ethical principles'' \citepalias{ArtificialIntelligenceForEurope}.

Due to the increase in computing power and the availability of large datasets, the performance of AI systems has significantly improved in recent years, and they are now routinely used in an ever-increasing number of applications ranging from biometrics and healthcare to the automotive domain. Despite the great opportunities offered by most recent AI systems, they also bring up new challenges, e.g., in the area of certification where auditing differs significantly from that of classical software systems \citep{berghoff2021towards, berghoff2022principles}.

\paragraph*{The current draft of the EU AI Act.}
The newly adopted draft of the EU AI Act \citep{EUAIAct23} aims to ensure that AI systems are overseen by people, are safe, transparent, traceable, non-discriminatory, and environment-friendly. The rules follow a risk-based approach and establish obligations for providers and users depending on the level of risk the AI can generate. Members of the European Parliament (MEP) substantially amended the list to include bans on intrusive and discriminatory uses of AI systems and expanded the classification of high-risk areas to include harm to people’s health, safety, fundamental rights, or the environment. MEP included obligations for providers of foundation models - a new and fast evolving development in the field of AI - who would have to guarantee robust protection of fundamental rights, health and safety and the environment, democracy and rule of law. They would need to assess and mitigate risks, comply with design, information and environmental requirements and register in the EU database \citepalias{EUAIAct21}.

\section{Guarantee of functional trustworthiness by risk-based development and testing}

By \textbf{functional trustworthiness} we understand all those aspects of trustworthiness that are directly dependent on the specific properties of the ML function itself,  which is the result of the data-depended optimization or learning algorithm.
It is understood that a final AI system consists of much more functionality surrounding the ML function, which is then usually developed using classical software development approaches. These surrounding parts of the application and the totality of the surrounding data and development managing system do of course have a great influence on the risk assessment and risk mitigation and thus on the trustworthiness of the final application.  In this section we want to highlight the differences between the classical and the data-driven programming approach in general and then specifically stress the necessary techniques for valid testing methods of the data-driven approach that is used in current AI systems. For the sake of clarity we restrict ourselves in this paper to discuss the ML-part and the ML-specific testing methods only, but we want to make clear that analyzing and testing the classical parts of the application with well-established classical certification procedures is of equal importance for the trustworthiness of the overall AI system.

\paragraph*{New paradigms through data-driven programming vs. classical software development.}
In the classical software development approach, humans design and write the functions and thus try to encode their knowledge to the best of their ability in the software with a clear formal goal in mind. The human designer takes the responsibility for the correct implementation of the function and in the ideal case, there is even a formal proof of correctness on some abstract level.
In contrast, in the data-driven approach a function is optimized according to a big corpus of data in view of certain technical optimization goals, which often differ for technical reasons from the intended performance measure of the application \footnote{e.g., classification accuracy vs. cross entropy loss}. The success of modern AI lies in the fact that the resulting functions and models have astonishing capabilities \footnote{generative models for photo-realistic images, large language models} and apparently work well on some samples at a first glance. It is thus tempting to believe in their magic and use these experimental models for various purposes. 

\paragraph*{Confidence by statistically valid testing.}
However this above described machine learning process by mathematical optimization alone gives no guarantees whatsoever, that the resulting function indeed fulfills the expectations in a specific application. For simple functions with just two or three parameters, such optimization processes are usually predictable and controllable. For the highly complex functions of deep learning models with millions or billions of parameters, however, the behavior of the optimization process and the resulting model is unpredictable. The optimization becomes a very high-dimensional pseudo-chaotic process and slight variations of hyperparameters or initialization of parameters could lead to very different final models \footnote{Astonishingly, it is often observed that most of these different models have very similar final performances, which shows the ambiguity of the optimization problem.}. 
On the one hand, we want to make use of the high complexity and expressiveness of sophisticated AI models, as it allows to extract and pick up detailed knowledge from all specific samples of the training data. On the other hand, that same property of expressiveness implies the uncontrollability of the learning process and its result. This implies the possibility that amidst of three or four input samples, where the function works properly, there may be another sample where it does not. 

Since the result of a learning process cannot be predicted analytically\footnote{except for low-dimensional or linear methods}, any relevant measure about the functional relation of the AI model with respect to the reality can only be expressed as a statistical measure.  
In order to evaluate a statistical measure we have to draw a number of independent random samples from the domain of the intended application, apply our learned function to them and count the number of good points where we are satisfied with the resulting behavior of the function versus the number of bad points where we are not. The ratio of good versus bad points is the simplest case of such a statistical measure. The same applies to any other characteristics of the function like fairness, gender bias, appropriateness of output statements, or any other measurable property\footnote{Note that specific inductive model biases may allow to shape the hypothesis space of models in favor of certain desired properties, like e.g. robustness. \citep{gehr2018ai2,cohen2019certified,jordan2021provable}. Furthermore there exist approaches for specific verification methods for neural networks, for those special cases where formal functional specifications can be given explicitly \citep{albarghouthi2021introduction}}. Under some mild conditions, we can fairly expect that this success/failure-ratio from a statistical  measurement result that is based on random sampling from the application domain will on average also apply to the future usage of this function. The decisive condition for the validity of this statistical measurement is the guarantee that no information whatsoever about those test samples can have leaked into the training process, neither via correlations with the training data, nor via any knowledge in the heads of the involved engineers who have designed the architecture of the ML function and the training process. These test samples have to be sampled independently and truly at random from the application domain. Any violation of this strict information separation is cheating and invalidates the results. 

\paragraph*{Well founded testing methods in ML theory.}
These testing principles and the necessity for a strictly separate, independently and randomly sampled testset are the theoretical foundations of machine learning, well known and accepted since the early days of this research field \citep{duda1973pattern}. We just have to stick to that well founded theoretical ground, even when it comes to the practical application.

\textit{So the crucial element of a correctly set up and trustable ML process is first and foremost a statistically valid testing procedure tailored precisely to the intended use case.}

In order to judge about the functional trustworthiness of an AI system we need:
\begin{enumerate}
    \item a precise definition of the application domain, formalized as a technical distribution that enables random sampling, 
    \item a definition of minimum performance requirements, that reflect the acceptable residual risks, and
    \item a statistically valid test of said performance requirements on independently sampled elements from the technical distribution.
\end{enumerate}


\paragraph*{The technical distribution of the application domain.}
It is understood that there is no mathematically formal way to define a probability distribution that reflects all possible samples and their probability of occurrence for any realistic application.
Therefore we have to resort to a process-like definition which describes how random samples are generated.
This implies a precise description of the intended use case. This must include all possibly relevant restrictions or conditions of samples for which the final application is designed. Such precise definitions have to include, e.g., the technical properties of sensors, like the resolution of digital cameras, the specification of the X-Ray machine, or simply the positioning commands for patients and their ethnicity for medical applications. Also, for regional or environmental specifications, a range or distribution has to be defined.
The technical distribution thus has to explain in words how one random sample at a time could (at least theoretically) be gathered, in such a way that the whole possible domain of the application is statistically covered. The verbal description has to be clear enough such that  users or engineers that are independent from the application development, but well informed about the application domain get to a consistent understanding of the distribution.  


\paragraph*{The definition of minimum performance requirements.}
Defining the minimum performance requirements is a most delicate task as it also involves the safety and ethical considerations for the application. The first challenge is to define suitable metrics, i.e., statistically measurable metrics, that can be applied on a totality of test samples. The second element is the definition of minimum values of these metrics that have to be met or surpassed by the AI system. These minimum values have to emerge from and be justified based on a risk-based analysis of the task. Performance metrics and corresponding minimum values have to be defined with regard to all risks of the application, especially in view of human safety and ethical values like non-discrimination, but also in view of the fairly expected behavior of the system while used by an average user. It is, e.g., fair to expect that a robotaxi will drive me in the direction of my goal, and not somewhere else. 

\paragraph*{Risk assessment determines minimum performance requirements.}
It is understood that a risk analysis and a risk management system have to be installed when an AI system is up to being brought to the market. 
Yet all hitherto regulation and standardization drafts fail to show and enforce that the risk assessment for the specific intended use of the application has to be the source of the required statistical quality of the AI model. 
Indeed, the authors want to stress here that the risk analysis has to be an integral part already in the development process of the ML system. 
It is precisely the risk analysis from which the statistical metrics and the minimum performance requirements have to emerge. 
In terms of the risk management, the likelihoods of those risks that emerge from undesired functional behavior of the ML-model are not a priori fixed given constants (as usual in classical risk management for engineering\footnote{see ISO/IEC 31000:2018 6.4.3 and ISO/IEC 23894 6.4.3.3}), but these likelihoods become variables, for which target values have to be defined. 
And it is precisely these target values of the risk likelihoods that constitute the minimum performance requirements for the ML-model. Consequently the model has to be tested in statistically valid tests precisely against those requirements. 

\paragraph*{Statistically valid testing.}  \textit{The technical distribution and the minimum performance requirements define a precise frame in which statistically valid testing of the ML system leads to an appropriate guarantee of functional trustworthiness.}

Based on the distribution and the metric, there is no doubt about how a statistically valid testing procedure is to be carried out. We have to generate a number of (independent) test samples from the above defined technical distribution of the application domain and measure the performance of the ML-model on those test samples with the above defined application-specific performance metrics. Based on the number of test samples and the measured results, we get a statistical confidence that the system fulfills the minimum performance requirements. This basic statistical testing procedure is simple and clear in its mathematical principle. It is worth noting that such a statistical testing procedure would be (a) impossible without a notion of the technical distribution, as independent random samples could not be generated, and (b) meaningless without minimum performance requirements, as no clear conclusion about the trustworthiness could be drawn from the test results.

\paragraph*{Big effort inevitable.}
The practical applications of this basic testing procedure may induce an overly big effort due to the necessity of a large number of independent testing samples. For many applications, this effort is inevitable, in order to achieve high confidence levels. Under certain circumstances, however, it is possible to resort in part to synthetic samples \citep{amodei2016concrete} or to deviate from the independence condition to a certain controlled extent. Further it may be advisable and efficient to reshape the distribution by drawing samples from critical regions more often than others and to adapt the performance measures accordingly. However, such alternatives and adaptations have to be supported by strong and precise mathematical arguments for the individual application under test. 


\paragraph*{The fallacy of fixed test sets.} It is a popular fallacy that something like fixed test sets could be used for the testing of AI applications. There even exists the misconception that fixed test datasets could be meticulously constructed as being specifically representative for certification purposes. This misconception seems to arise from the concept of benchmark datasets. It is true that fixed public benchmark datasets are commonly used in scientific machine learning research in order to provide a well defined fixed synthetic environment for comparing properties of different architectures and training algorithms, often also under the restriction of a fixed training dataset, as part of the benchmark definition. The scientific goal is to benchmark the individual algorithmic tweaks and tricks, analyzing their impact, and generating general methodological insight. Notably, in the context of benchmark datasets it is the methodological approach of ML that is tested, and not the applicability of a specific ML model to the real world.
Outside of academic research, however, the job of a machine learning application developer consists in the engineering of a specific ML model that performs sufficiently well in the real world -- with respect to the application-specific metrics. Using weak and tricks for constructing the training dataset is perfectly allowed here with the intention to achieve better final performances. Even manual training dataset construction to include specific cases is a valid and widely used practice. As soon as there would be a fixed test dataset for the commercially important purpose of certification, it could never be ruled out that some information from such a fixed test dataset would find its way into the training process. Such information leakage would then invalidate the measurement results \citep{kaufman2012leakage}. 

\paragraph*{The fallacy of training set regulations.}
It is a popular belief that fairness and non-discrimination of the resulting model could be guaranteed by enforcing that the training set shall be "representative" of the application domain. This is just a fallacy and not true. Even if the functional model itself would give precisely the same support to all training samples -- which by itself is very improbable -- there are no means to ensure that every training sample will have an equal effect on the optimization process \citep{meding2021trivial}. Such equal weighting is not at all the goal of the training process. Even though it is often true that the mathematical procedure of the training process attributes equal importance to all training samples, the effect of the optimization is dominated by the spurious higher order nonlinear correlations between the samples. Implicitly finding these highly intricate abstract correlations is precisely the root of the emerging "intelligence" of the trained model. Furthermore, there are even very powerful optimization algorithms (e.g., support vector machines) that make heavy use of dynamic re-weighting of the training samples in order to achieve the best optimization results with respect to the goal function. 
Any functional requirement to an ML model, be it correctness, robustness, desired or undesired biases, fairness or non-discrimination, can only be assessed by statistically valid tests on the trained final model.  
That said, it is understood that the training set is an essential element for the development of ML models. It is the duty of the involved developers of an AI system to design and tweak the training set, to create learning schedules with different subsets, to augment the training set in various ways even beyond the technical distribution of the application, or to add artificially created samples, as long as all those tweaks serve the purpose of achieving the desired behavior of the trained function. 
\textbf{Any regulation that directly restricts the freedom of design of the training set\footnote{Just for the sake of clarity, note that lawful data accessibility restrictions and lawful transparency obligation are of course not part of the free design decision.} is at best futile and at worst counterproductive. Propagating the illusion that such regulations would positively contribute to the trustworthiness of AI is at best naive and at worst dangerous.}

\paragraph*{Conclusion: Functional trustworthiness by risk-based development and statistical facts.} The grounding of any judgment about the trustworthiness of an AI/ML system consists in the clear and credible documentation of the three elements, (1) definition of the technical distribution, (2) risk-based minimum performance requirements, and (3) valid testing based on independent random samples. Together with a number of other quantitative and qualitative tests, these are three key-elements should be the necessary foundation for fact-based decisions about the certification of an AI system and its admission to the market.

\section{Further requirements for trustworthy AI}
The above description of a risk-based AI model development only covers the principles of the functional aspects of trustworthiness. As mentioned above there are of course many other relevant details and further aspects to the trustworthiness of an AI system. In the following we will highlight some of those aspects, especially those that currently lead to great confusion and discussions both in the standardization committees and in the draft of the EU AI Act, but also in the general public.

\paragraph*{Locked vs. Adaptive AI systems.}
From the point of view of standardization and certification, it makes sense to discriminate between at least three different kinds of AI systems:
\begin{itemize}
    \item \textbf{Locked AI:} This is the basic form of the above described data-driven programming. The once developed AI model is finally assessed and its functional behavior is not further changed by the repeated application in the field. From time to time, an updated version of that AI system may be put on the market, which then has to be assessed again.
    \item \textbf{Updated AI:} This extends the basic form of locked AI by semi-automatic or fully-automatic - yet human supervised - regular updated versions of a further optimized AI model\footnote{\href{https://www.fda.gov/files/medical\%20devices/published/US-FDA-artificial-Intelligence-and-Machine-Learning-Discussion-Paper.pdf}{Proposed Regulatory Framework for Modifications to Artificial Intelligence/Machine Learning (AI/ML)-Based Software as a Medical Device (SaMD) - Discussion Paper and Request for Feedback}}. Due to the fast pace of that automated update cycle, the iterated model versions cannot be tested with the same rigor as the single initially trained and assessed model. This is especially problematic, as every change of the function can lead to both positive as well as negative impacts with respect to the minimum performance requirement of the initial assessment. The true effect of any change can only be measured by a new statistically valid test.
    \item \textbf{Adaptive, self-optimizing AI (e.g. self-supervised learning, reinforcement learning):} Such AI systems are capable of adapting themselves to new environments or finding solutions to new, hitherto unsolved problems. The evolution of the function of such a self-optimizing AI system cannot be known or tested in advance, neither by statistical tests, nor by any other functional assessment. Its evolving behavior is defined only by some goal function and its internal capabilities. In a restricted environment or in simulated virtual realities these AI methods have demonstrated their unique capability of creating originally new knowledge \footnote{Deepmind AlphaFold, Tokamak plasma control, Alpha Go Zero etc.}. Due to the complexity of the unrestricted real-world, it is impossible to predict the behavior that emerges from a certain goal function \footnote{paper-clip problem}. At the moment of writing this paper, nobody does seriously consider to release such self-optimizing AI into the wild.
\end{itemize}
It is understood that the above described approach of risk-based development and testing of AI focuses primarily on the assessment of locked AI systems. Yet, under certain conditions, it should be possible to adapt the above procedure such that the minimum performance requirements will still be fulfilled after a certain small amount of controlled updates, after which a new assessment is advisable.

\paragraph*{Alignment and the fallacy of AI dystopias.}
It is worth noting that all apocalyptic endtime AI dystopias of careless or malevolent AI that wipe mankind from earth only apply to the last case of adaptive, self-optimizing AI. All currently applied AI systems do not fall into this self-optimizing category. The so called \textit{alignment problem} \citep{wiener1960some} addresses in its most general form the currently unsolvable problem, how to guarantee in advance that a certain set of goal functions of a self-optimizing AI system indeed leads to the emergence of the desired human-friendly behavior. Solving this most general form of the alignment problem would require that the above described certification of functional trustworthiness, including the risk analysis, would have to be handed over to a fully automated process. We are far from being able to even think about that.
The fact that the EU AI Act does not clearly distinguish between the three categories of AI systems (locked, updated, self-optimizing) is probably one of the biggest shortcomings of this draft\footnote{apart from the omission of any regulations of military applications}.



\paragraph*{Adversarial Attacks (Security).}
Adversarial attacks still pose a threat to the safety of ML-systems.
Such attacks are intentionally designed and aim to manipulate the inputs to the ML-systems in order to force them to differ from their expected behavior \citep{szegedy2013intriguing}. 
This can lead to critical failures and hazardous situations. The key aspect of adversarial samples is the fact, that they usually are not samples from the intended target distribution of the application. Thus, the statistical testing approach alone, which samples randomly from the target distribution, is not sufficient to measure the risks of malevolent intentional adversarial attacks. In addition to the ML aspect, classical software development security issues also play a major role in the defense of such attacks and should prevent the direct accessibility of the ML-model for adversarial data as far as possible. 

\paragraph*{Robustness, Domain shifts, Out-of-distribution detection.}
Depending on the kind of application and the possible risks it might be necessary to test the final ML-model also with further samples that are outside of the technical distribution of the application domain. The term \textit{robustness} is sometimes used to refer to the quality of the modeled function with respect to such samples. The definition of statistically valid measures for the various arising risks are often very difficult and sometimes impossible to design \citep{david2010impossibility}.
An unsolved problem regarding certification arises from domain gaps (also called distributional shifts) \citep{quinonero2008dataset, david2010impossibility, zellinger2021generalization, fischer2020ai}. 
Domain gaps arise when the data distributions for learning and inference do not match, which can be quite common for real-world applications. 
A typical scenario is learning from simulated data and 
applying the model in the real world which usually differs at least slightly from the simulation. It is understood that statistical valid testing for functional trustworthiness relies on real-world data. 

\paragraph*{Assistant AI systems and human oversight.}
By \textit{assistant AI systems}, we refer to systems that do not by themselves exercise decisions or act on the real-world environment but just prepare data, analyze information, and propose decisions or generate images \citep{goodfellow2014generative,ho2020denoising} or texts \citep{vaswani2017attention,Schulman_Zoph_Kim_2022}. The output of the system is thus targeted at a competent user who has the authority to accept, reject or modify the suggested result of the AI system. This collaboration between the expert knowledge of the AI system and the authority of the well-informed or not so well-informed user has a very complex dynamics that is usually characterized by mistrust in the beginning of the collaboration and an overconfidence after a certain time of use.

In order to assess the risks of such assistant AI systems, it is of decisive importance that a clear profile of the intended user of the system is defined. 
Common navigation systems for cars, e.g., may assume that the user, i.e., the driver of the vehicle, is an experienced driver and able to judge if the proposed path is legally and physically viable. 
The responsibility for the specific chosen path remains 100\% at the driver. An assistant for the detection and classification of mushrooms, however, will probably not be intended as a tool for botany experts, but addresses primarily uninformed users that might rely on the recommendation of the system for which mushrooms they can eat or not. 
A crucial part of the human-AI collaboration is the information of the user about the system and its performance metrics as this information shapes the expectations of the user.
This will be especially important for medical AI assistants. In order for such systems to be practically useful, certain guarantees and responsibilities for the quality of the system have to be imposed on the one who puts the system on the market. But the responsibility to judge if the individual case falls into the defined regime of the system and if the results are plausible in the larger context of the whole medical case will remain at the physician that uses the system. Other ethically critical domains for such assistant AI system are the pre-selection of job applications or human resources decisions in general, or insurance ratings and credit worthiness judgments. 

The key for an effective human supervision is thus the transparency of the system with respect to the three objectives of the functional trustworthiness: the precisely defined distribution of the intended use cases, the calculated risks or side-effects as per the risk-based minimum performance requirements, and the resulting test performance of the statistical tests. These parts of the technical documentations have to be accessible to all users and have to take sufficiently into account the risk of overconfidence with regard to the risks and critical effects on the user.

\section{Personal AI assistants}
Personal AI assistants have gained an enormous popularity in these last months since the release of ChatGPT by OpenAI in November 2022. Thus, these systems are an important subclass of AI systems and we would like to discuss the applicability of the principle of functional trustworthiness to this subclass in this section. Personal AI assistants certainly deserve a special consideration with respect to regulation, standardization, and certification, thus we dare to pinpoint also some of these currently discussed aspects that go beyond testing and functional trustworthiness.

The basis of these personal AI assistants are generative models.
Their variation ranges from pure text-based chatbots \citep{Schulman_Zoph_Kim_2022} over text to image generators \citep{ramesh2021zero} to combined systems that use video, audio, music, and text for both input and output \citep{reed2022generalist}. The special characteristics of these systems is that their application domain covers a very broad range of fields, including science, arts, politics, jurisdiction, and even the automatic generation of scripts and program code. They are primarily intended to support the users in their personal creative process, but at the same time they encode so much human knowledge that it is tempting to take the superficially plausible output as true factual statements.

\paragraph*{Functional trustworthiness -- Typical prompting.} 
In the current usual application scenario of a personal AI assistant, a more or less knowledgeable human being uses such a system in order to generate new ideas and textual, graphical, or audible representations of them. These ideas and their representations might be used for pure recreational purposes, for personal philosophical interests, for work-related duties, for the support of scientific research, or for political activities and other interests at the discretion of the user.
The above presented three elements for establishing a functional trustworthiness can in principle be applied to AI assistants, even if they consist in large language models or generative ML models as described in the last section. Thus, it is perfectly possible to create empirical tests for such models. It can be assessed whether the system answers factually correct to typical questions or if the descriptions or images that result from typical prompts correspond to certain minimum performance requirements, e.g., the desired degree of non-discrimination. Thus it is perfectly possible to test the application scenario of the response upon typical first prompts. This restricted application scenario is a potential basis for a certifiable  application domain. 

\paragraph{Functional trustworthiness -- Adversarial prompting.} Yet for interactive conversational systems, as e.g., ChatGPT \citep{Schulman_Zoph_Kim_2022} and its numerous variants \citep{thoppilan2022lamda, scao2022bloom, kopf2023openassistant, taori2023alpaca}, there is a crucial difference to all other domain-specific and knowledge-specific 
applications in the sense that there is no necessity for an objective, external input as compared to special purpose narrow AI applications. In medical applications, e.g., there are more or less well defined inputs from measurements from a patient. In door-opening systems, there is the recorded image or video or voice input, which is the basis of certain well defined decisions. In self-driving applications, there are the current measurements of the environment, that represent the input data. In personal AI assistants, however, the inputs, i.e., the so called prompts, are generated arbitrarily by the human users, at their sole discretion. And if the users are not satisfied with the result, then they will modify this input or add a second prompt to the session in order to achieve their goal. Overall, the typical application case is a presumably "intelligent" user who performs a directed search on the highly complex ML-function of the assistant. While it is perfectly measurable with the methods of functional trustworthiness to analyze the responses to typical first prompts, it is nearly impossible to predict, which other responses might possibly be generated if the user explicitly and intentionally searches for them. This active search for a specific output is precisely the setting of adversarial attacks, where the user, with benevolent or malevolent intentions, is the attacker of the system. There are numerous reports about such successful "attacks", e.g., against the ethical rules of ChatGPT, by setting a possibly censored question in a hypothetical context ("imagine you are a ... system.") or by asking indirectly ("what would person x say, if he was asked ...") \citep{wang2023robustness}. As already explained earlier, a statistical performance measure, i.e., a test based on randomly sampled prompts, is not predictive for adversarial robustness.
In other words, as long as the surrounding system does not simply cut off the conversation upon the more or less simple recognition of certain words or topics, \footnote{compare with the false recognition of Michelangelo's David as porn}  there will always be ways to generate undesired output by clever prompt engineering.

\paragraph{Output restrictions.}
We appreciate the voluntary effort that OpenAI and other providers of personal AI assistants are currently investing in order to align the current systems with ethical values. It is worth noting that this effort is not governed by legal restrictions, but by the goal of the providers themselves to make their systems as attractive as possible to a large community of users. However, there are two notable downsides of that approach. 

First, it has been empirically observed, that such a restricted system has a lower quality of creativity and is less capable of abstract far-fetched associations. And this reduction in quality does not only concern the specific topic field where restrictions are intended, but due to the generalization capabilities, the restrictions also have uncontrollable effects on other topics. Obviously there exists a delicate balance between desirable restrictions and the generative quality of the system. As explained earlier, due to the adversarial nature of the user, there are no clear methods to measure if the system's output restrictions are sufficient or not. Legal restrictions on the other hand should be clear and executable, so as to create a solid ground of rules for economic players and the society. The basic requirement that could be legally imposed is the restriction that the system should not present \textit{unwanted} unethical outputs, i.e., the system should not by itself drift away from the requested topic and bother the user with unethical imperatives. This kind of requirement excludes the risk of a possibly adversarial objective of the user, and thus this requirement can perfectly be tested for by the above statistical tests of functional trustworthiness.

Second, every content restriction is also a certain censorship, that restricts the freedom of thought-building of the concerned user. Most hitherto drafts for the regulation of AI, as e.g., the EU AI Act, are intended to regulate \textit{automated decision making} in certain critical fields or in view of security risks. Restrictions on the admissible content of private AI assistants, however, is a regulation of \textit{ideas and thought generation}. 
A community-wide legally imposed restriction that excludes certain ideas or topics from all these personal AI assistants would be a very efficient censorship of thoughts. This would be much more efficient than any historical censoring tool or censoring process, as it would affect already the seed of private personal opinion building, not only its spreading.
Such fundamental restrictions of personal freedom should therefore be balanced very carefully against the intended benefit. 

\paragraph{Risk of possible deadlock by censorship.} It is to be assumed that in the not-too-distant future, everyone will interact with a personal AI assistant for their daily work, probably even in a much more seamless way as compared to how we interact with our smartphones today. If we imagine such a world all the way down to its end, this would have remarkable, probably unwanted side-effects. In a society where all documents and thus also all political statements would be prepared using those restricted personal AI assistants, it would be very difficult or nearly impossible to have an open public democratic discussion and opinion-building about possible changes of the censoring regulations.


\paragraph{Fallacy that AI spams us with fake-news.}
Fake-news are definitely one of the current problems of our digital society. Yet fake-news, lies and rumors are as old as mankind and language \footnote{There are even reports about some species of animals that falsely yell danger to its fellows, if they have secretly found a new source of food hitherto unknown to those fellows.}. 
It is of course true that personal AI assistants could be misused in order to generate fake-news, deep-fake images and convincingly written false texts.
One of the signs that we have hitherto used in order to identify maliciously faked texts or messages was the quality of the text itself. 
Intuitively we have more trust into a text if it is consistent in itself and grammatically correct, simply because we have more trust in educated people than in uneducated people. At its base, we have to realize that this is just a prejudice, as the education of persons and their intention not to lie to our face are not related to each other. Nevertheless, in the last decades the world-wide internet users got used to take the formal language or image quality of the message as a proxy indicator of the quality and trustworthiness of the content.
It is obvious that this was already a bad proxy in the past, and it is clear that in view of deep generative models and deep-fakes this is a horrible proxy for the future. Yet, we doubt that regulating AI will really help against that phenomenon\footnote{Big techs will probably have to develop suitable methods on their own behalf in order to avoid degeneration of models by accidentally learning from generated content \citep{shumailov2023curse}}. The obligation to include watermarks or similar indicators into AI-generated output could possibly help to discover cheating pupils in schools, but really malevolent actors will have the tools to remove such watermarks. Furthermore, malevolent providers of AI models in the darknet will just ignore such regulations all together.

\paragraph{Reducing spread of fake-content by cryptographic signatures.}
There is just one thing that even the most powerful AI-models will never be able to generate or fake\footnote{not even future AI systems, as long as the length of the key is sufficiently large  and as long as real quantum computers do not exist}: cryptographic keys and signatures.
The technology is already there for decades \citep{rivest1978method} and widely used in IT networks in order to establish trust relationships, e.g., between user accounts and operating systems and services. 
For certain applications and documents, cryptographic signatures are already in use. The PDF format actively supports signing PDF documents. Also some email applications support cryptographic signatures and check the signature of the sender of incoming emails. It is only very little effort that would be necessary in order to create a global web of trust and the possibility of fully transparent signed data flows. This should e.g. include that every download is signed by the respective website. Images should be signed by the real physical camera that took the picture or by the image processing tool that signs the processed result and that may also includes the info or hash-key about the original source of the image. It could still be at the discretion of the authors of messages whether to disclose their sources of information to the recipients, but if all sources are signed, the authors at least have verifiable proofs to do so.
Not including the original sources would probably decrease the credibility of the message in the eyes of the recipients, thus providing sources with signatures that can easily be checked by everyone
would benefit the sender in the eyes of the community.

For certain reasons that we do not analyze here, signatures are typically not in use in social media, and only very few users apply them for their daily emails. Anonymous social media and email accounts are much more widespread than users that are identified by name. It is understandable that users lost the intuition of being responsible for their messages if they can hide their identity. Yet anonymity also has its value for freedom and should not be abolished or forbidden. However, it might be considered that only those messages that are cryptographically signed can surpass a certain spread or go viral. The sender of signed messages could also have a higher legal responsibility for the content, maybe similar to registered news papers today. Obviously, a valid signature would have to contain the necessary information to start legal actions against the content, without the necessity to ask the provider of the platform to disclose the user account information.
It is clear that a regulation of social media, blog posts and similar widespread publication platforms will by itself create a lot of legal questions. And fostering the acceptance of a wide spread web of trust, that should notably be independent of the providers of the social media platforms, is certainly not trivial. But if the spread of fake information is identified as the root of a problem it seems reasonable at least to consider such more effective ways instead of prolonging the problem with inefficient regulations for AI.

\paragraph{Responsibility of the user.}
It remains the risk that the users are unaware of their responsibility, place a false trust in the texts and ideas generated, are still minors and need to be protected, or have limited legal responsibility for their actions for other reasons. These risks have to be mitigated appropriately by the surrounding application, suitable warnings, and access restrictions. Also, the appearance of additional topic-specific warnings or legal hints might be considered, but should not interfere with the free generation of results.


\paragraph*{Further risks.} 
There are of course a number of further risks to observe, when such generative AI models are designed, developed, and brought to the market (e.g., privacy, copyrights of training data, authorship of generated data, transparency of training data). We will not elaborate on these risks in this document, as they are not elements of the scope of functional trustworthiness, i.e., they cannot and should not be aimed to be imposed or checked based on statistical tests on the final model. Of course, these topics are crucial parts of other, additive audit sections. Their mitigation lies in other parts of the system: privacy risks have to be covered by high quality data management both during the development and at run time; copyright questions have to be taken into account at the selection of the training data; and the legality of authorship is a questions for the legislator.

\section{Conclusion and recommendations}
In collaboration with the Johannes Kepler University Linz and the Software Competence Center Hagenberg, TÜV AUSTRIA has developed an elaborated catalog for the certification of AI systems \citep{winter2021trusted} on the basis of the current state of the art in ML science and gathered extensive experience of its application in a number of pilot projects. At the time of writing, this is the first and hitherto only commercially available certificate of trustworthiness for AI systems that does not only certify the development process, but also the quality of the final product.
In this paper we have shown that the assessment of the trustworthiness of any AI system has to include at its heart the assessment of the functional trustworthiness, that consists in
\vspace{-0.3cm}
\begin{itemize}[noitemsep]
    \item[\textbf{(1)}] the definition of the technical distribution of the application domain,
    \item[\textbf{(2)}] the risk-based definition of minimum performance requirements, and 
    \item[\textbf{(3)}] the statistically valid testing of the final models based on independent random samples.
\end{itemize}
In fact, currently most of the serious commercial providers of AI software are well aware of the necessity of functional trustworthiness. They at least implicitly strive to follow the described principles in their current development pipelines and product cycles.
 On the other hand, we see time and again in projects that the complexity of implementing and documenting the details of these principles in a transparent and certifiable way challenges even advanced developers to the limits of their capabilities. We would thus like to invite all persons that are currently involved in the regulation and standardization of AI systems to join us in advocating for harmonized rules and standards that put the functional trustworthiness and its transparency at the core of the conformity requirements.

\vspace{2cm}

\section{Acknowledgment Statements}

This work has received funding from the European Union’s Horizon 2020 research and innovation program under grant agreement No 951847. The project ELISE works in close collaboration with the ELLIS Society (European Laboratory for Learning and Intelligent Systems).

The research reported in this paper has been funded by the State of Upper Austria within the strategic program \#upperVision2030 (Certification of AI, Phase B) and the Federal Ministry for Climate Action, Environment, Energy, Mobility, Innovation and Technology (BMK), the Federal Ministry for Digital and Economic Affairs (BMDW), and the State of Upper Austria in the frame of SCCH, a center in the COMET - Competence Centers for Excellent Technologies Program managed by Austrian Research Promotion Agency FFG.

We would like to extend our sincere appreciation to our reviewers for their valuable feedback and contributions, which significantly improved the quality of this paper.

\bibliographystyle{plainnat}
\bibliography{literature}

\newpage
\section{Reviewer Statements}

\paragraph*{Reviewer A, short statement:} \textit{While the EU AI Act is seen as an important step towards regulating artificial intelligence (AI), it also has certain limitations that should be acknowledged. Its broad scope and stringent requirements can be challenging for both AI developers and users, potentially hindering innovation and the development of useful AI applications. In addition, the Act's provisions may not address the rapid pace of AI advances, making it difficult to keep pace with evolving technologies and emerging risks.}
 
\textit{Developing trustworthy AI systems is a complex task due to several factors. It requires a multi-layered approach that includes robust and transparent technical methodologies, as well as comprehensive testing and governance frameworks. This position paper highlights the shortcomings of AI Act in this regard, particularly with respect to ensuring functional trustworthiness. The authors provide valuable insights into why and how a systematic testing regime for AI systems should be implemented.}

\textit{They make the case for application-specific testing of AI systems, as this is critical to ensuring their reliability and safety. Different AI applications have unique requirements and potential risks. Conducting thorough, application-specific testing and validation can help identify vulnerabilities, eliminate biases, and mitigate potential damage. Tailored testing approaches are essential to understand the limitations and capabilities of AI systems, promote responsible use, and build trust with users and stakeholders.}
\vspace{0.2cm}
\begin{quote}
\textbf{Prof. Dr.-Ing. Marco Huber} \\
Head of Center for Cyber Cognitive Intelligence CCI \\
Head of Department Image and Signal Processing \\
Head of AI Innovation Center \\
Fraunhofer Institute for Manufacturing Engineering and Automation,\\ Stuttgart, Germany \\
\end{quote}

\paragraph*{Reviewer B, short statement: } \textit{This article raises the important question of whether the current draft texts of the AI Act are too vague with respect to quality requirements concerning
the development of data-driven AI systems with machine learning techniques.  It is excellent that this issue is brought up at this very moment, since without proper contemplation now, this may later result in poorly formulated conformity assessment and standardization procedures that do not properly support the true spirit of the AI Act, and may not provide adequate protection for the safety and health of people and for their ethical values and fundamental rights. All clarifying principles, best practises or supportive tools leading the regulation and standardization activities in the right direction are more than welcome, and this article is a great contribution in this process.
}
\vspace{0.2cm}
\begin{quote}
\textbf{Prof. Petri Myllymäki, Ph.D.}\\
Artificial intelligence and machine learning\\
Department of Computer Science, University of Helsinki, Finland\\
Director, Helsinki Institute for Information Technology (HIIT)\\
Vice-Director, Finnish Center for Artificial Intelligence (FCAI)\\
Director, Helsinki Doctoral Education Network in ICT (HICT)\\
\end{quote}

\paragraph{Reviewer C}wants to stay anonymous. He is a professor at a well-renown research center in Germany and Fellow of the ELLIS-Society.

\end{document}